\documentclass[11pt]{article}

\usepackage[utf8]{inputenc}       
\usepackage[T1]{fontenc}
\usepackage{lmodern}
\usepackage{microtype}

\usepackage{amsmath,amssymb,amsfonts,amsthm,mathtools}
\usepackage{booktabs}
\usepackage{authblk}
\usepackage[margin=1in]{geometry}
\usepackage{xcolor}
\usepackage{hyperref}
\usepackage{enumitem}
\usepackage{listings}
\usepackage{caption}

\hypersetup{
	colorlinks=true,
	linkcolor=blue,citecolor=blue,urlcolor=blue
}

\newcommand{\Xp}{\mathbf{X}_{p}}
\newcommand{\Hp}{\mathbf{H}_{p}}
\newcommand{\Yp}{\mathbf{Y}_{p}}
\newcommand{\W}{\mathbf{W}}

\newcommand{\R}{\mathbb{R}}
\newcommand{\cond}{\kappa}
\newcommand{\norm}[1]{\left\lVert #1 \right\rVert}

\newcommand{\barsig}[1]{\bigl[\mathbf{1},\,\sigma(#1)\bigr]}  
\DeclareMathOperator{\softplus}{softplus}
\DeclareMathOperator{\Tr}{Tr}

\newtheorem{prop}{Proposition}
\newtheorem{lemma}{Lemma}

\title{\bf Prototype Training with Dual Pseudo-Inverse and Optimized Hidden Activations}
\author{Mauro Tucci}
\affil{Department of Energy and Systems Engineering, University of Pisa, Italy}
\date{}

\begin{document}
	\maketitle
	
	\begin{abstract}
		We present Proto-PINV+H, a fast training paradigm that combines closed-form weight computation with gradient-based optimisation of a small set of synthetic inputs, soft labels, and—crucially—hidden activations. At each iteration we recompute all weight matrices in closed form via two (or more) ridge-regularised pseudo-inverse solves, while updating only the prototypes with Adam. The trainable degrees of freedom are thus shifted from weight space to data/activation space. On MNIST (60k train, 10k test) and Fashion-MNIST (60k train, 10k test), our method reaches \textbf{97.8\%} and \textbf{89.3\%} \emph{test accuracy on the official 10k test sets}, respectively, in \textbf{3.9--4.5\,s} using approximately 130k trainable parameters and only 250 epochs on an RTX 5060 (16\,GB). We provide a multi-layer extension (optimised activations at each hidden stage), learnable ridge parameters, optional PCA/PLS projections, and theory linking the condition number of prototype matrices to generalisation. The approach yields favourable accuracy--speed--size trade-offs against ELM, random-feature ridge, and shallow MLPs trained by back-propagation.
	\end{abstract}

	\section{Introduction}
	Back-propagation (BP) is the de facto standard for neural network training but requires full-graph gradient computation and storage of intermediate activations, which increases training time and memory usage. Analytic or semi-analytic methods (e.g. Extreme Learning Machines) reduce runtime by solving linear systems but typically forgo accuracy on modern benchmarks.
	
	\paragraph{Key idea.}
	We relocate trainable degrees of freedom from the weight space to a compact set of \emph{prototypes}:
	synthetic inputs \(\Xp\), hidden activations \(\Hp\), and soft labels \(\Yp\).
	At every iteration, we analytically recompute the weights via ridge-regularised pseudo-inverses, and only update \((\Xp,\Hp,\Yp)\) with gradient descent. Optimising \emph{hidden activations} is a central novelty that improves expressive power beyond methods that optimise only input prototypes and labels.
	
	\paragraph{Contributions.}
	\begin{itemize}[leftmargin=1.25em]
		\item A dual pseudo-inverse procedure with \emph{optimised hidden activations} (Proto-PINV+H), enabling closed-form weights at every step.
		\item A multi-layer extension with analytic inter-layer weights and learnable prototype activations at each hidden stage.
		\item Learnable ridge parameters and optional PCA/PLS projections; analysis of numerical stability and complexity independent of dataset size.
		\item A theoretical connection between prototype matrix conditioning and generalisation, plus differentiability of the ridge solutions.
		\item Strong accuracy--speed--size trade-offs on MNIST and Fashion-MNIST with 250-epoch runs on a single consumer GPU (RTX 5060, 16 GB).
	\end{itemize}
	
	\section{Background and Preliminaries}
	\subsection{Ridge-regularised least squares}
	For a design matrix \(A\in\R^{m\times n}\) and targets \(B\in\R^{m\times r}\), ridge-regularised least squares solves
	\[
	\min_{W\in\R^{n\times r}} \ \frac{1}{2}\norm{AW-B}_F^2 + \frac{\lambda}{2}\norm{W}_F^2,
	\]
	whose solution is \(W=(A^\top A+\lambda I)^{-1}A^\top B\). This is numerically stable for \(\lambda>0\) and differentiable with respect to \(A,B,\lambda\) via the implicit function theorem. In practice we compute \(W\) by solving \((A^\top A+\lambda I)W=A^\top B\) with a Cholesky-based solver.
	
	\subsection{Differentiability through linear solves}
	Let \(G(\theta)=A(\theta)^\top A(\theta)+\lambda(\theta)I\) and \(R(\theta)=A(\theta)^\top B(\theta)\).
	Then \(W(\theta)=G(\theta)^{-1}R(\theta)\). For a scalar loss \(L(W)\),
	\[
	\frac{\partial L}{\partial \theta}
	= -\Tr\!\bigl((G^{-1}R\,\nabla_W L^\top) \, \partial_\theta G\bigr)
	+ \Tr\!\bigl(G^{-1} \partial_\theta R \, \nabla_W L^\top \bigr),
	\]
	where \(\nabla_W L\) is supplied by reverse-mode AD. Modern autodiff libraries expose this without manual Jacobians.
	
	\section{Method}
	\subsection{Setup and variables}
	Given labelled data \(\{(x_i,y_i)\}_{i=1}^N\) with \(x_i\in\R^d\), \(y_i\in\{1,\ldots,k\}\),
	we synthesise \(N_p\ll N\) prototypes:
	\[
	\Xp\in\R^{N_p\times (d+1)},\quad
	\Hp\in\R^{N_p\times h},\quad
	\Yp\in\R^{N_p\times k},
	\]
	where \(\Xp\) includes a bias column and \(\barsig{\Hp}=[\mathbf{1},\sigma(\Hp)]\) augments the nonlinearity with a bias column.
	
	\subsection{Dual pseudo-inverse with optimised activations}
	\paragraph{Closed-form weights.}
	For \(\lambda_1,\lambda_2>0\),
	\begin{align}
		\label{eq:w1}
		\W_1 &= (\Xp^\top\Xp + \lambda_1 I)^{-1}\Xp^\top \Hp,\\[0.6ex]
		\label{eq:w2}
		\W_2 &= \bigl(\barsig{\Hp}^{\!\top}\,\barsig{\Hp} + \lambda_2 I\bigr)^{-1}\barsig{\Hp}^{\!\top}\,\Yp.
	\end{align}
	Given a real input \(x\), predictions are
	\[
	f_\theta(x) = \barsig{\,\sigma\bigl([\mathbf{1},x]\W_1\bigr)\,}\,\W_2.
	\]
	
	\paragraph{Objective.}
	Let \(F=[\mathbf{1},X]\) be the bias-augmented training data. We minimise
	\begin{equation}
		\label{eq:loss}
		\mathcal{L} = \mathrm{CE}\bigl(\barsig{\,\sigma(F\W_1)\,}\W_2,\ y\bigr)
		+ \lambda_3 \bigl(\norm{\W_1}_F^2 + \norm{\W_2}_F^2\bigr),
	\end{equation}
	updating only \(\Xp,\Hp,\Yp\) (and optionally \(\lambda_1,\lambda_2\) via \(\lambda=\softplus(\rho)\)).
	
	\subsection{Multi-layer extension}
	For depths \(L\ge 1\) with widths \(h_1,\ldots,h_L\), introduce prototype activations \(\Hp^{(l)}\) and define
	\[
	\W_1=\Xp^+\Hp^{(1)},\quad
	\W_2=\barsig{\Hp^{(1)}}^{\!+}\Hp^{(2)},\ \ldots,\ 
	\W_{L+1}=\barsig{\Hp^{(L)}}^{\!+}\Yp,
	\]
	where \(A^+=(A^\top A+\lambda I)^{-1}A^\top\). Only \(\Xp,\Hp^{(l)},\Yp\) are trainable; all \(\W_l\) are recomputed in closed form.
	
	\subsection{Regularisation, soft labels, and initialisation}
	\textbf{Weight penalties.} We use an explicit Frobenius penalty on \(\W_1,\W_2\) via \(\lambda_3\), distinct from ridge terms \(\lambda_1,\lambda_2\).  
	\textbf{Soft labels.} \(\Yp\) is optimised directly; optionally a temperature \(T>0\) enforces \(\tilde{Y}_p=\mathrm{softmax}(\Yp/T)\).  
	\textbf{Initialisation.} Balanced one-hot on \(\Yp\) often accelerates early training; \(\Xp\) can be random normal or stratified from training samples. \(\Hp\) is random normal.  
	\textbf{Decoupled weight decay.} Separate decays for \(\Xp\) and \(\Hp\) give finer control; decay on \(\Yp\) is often unnecessary with soft labels.
	
	\subsection{Dimensionality reduction (optional)}
	PCA or PLS projects \(x\mapsto U^\top(x-\mu)\) to \(d'\ll d\) before prototype optimisation. After training, we back-project \(\W_1\) to original pixels; the mean shift is absorbed by the bias.
	
	\subsection{Complexity and stability}
	Each epoch solves small ridge systems on \(N_p\) prototypes and backpropagates only through \(\Xp,\Hp,\Yp\). The cost scales as \(\mathcal{O}(N_p h^2)\) plus \(\mathcal{O}(h^3)\) for linear solves, independent of \(N\). Cholesky-based solvers with positive \(\lambda\) ensure numerical stability; SVD can be used as a fallback in highly rank-deficient regimes.
	
	\section{Theory}
	\subsection{Stability of ridge solutions and conditioning}
	\begin{lemma}[Ridge sensitivity]
		Let \(W_1(\Xp,\Hp)=(\Xp^\top\Xp+\lambda_1 I)^{-1}\Xp^\top \Hp\).
		Then, for small perturbations \(\Delta \Xp,\Delta \Hp\),
		\[
		\norm{\Delta W_1}_F \ \le\ 
		C_1(\lambda_1,\Xp,\Hp)\,\norm{\Delta \Xp}_F
		+ C_2(\lambda_1,\Xp)\,\norm{\Delta \Hp}_F,
		\]
		where \(C_1\) scales with \(\sigma_{\max}(\Xp)/\sigma_{\min}(\Xp)\) and \(C_2\) with \(1/\sigma_{\min}(\Xp)\).
	\end{lemma}
	
	\begin{prop}[Generalisation via condition number]
		\label{prop:cond}
		Assume the loss is 1-Lipschitz in \(\W_1\). Then, with probability at least \(1-\delta\),
		\[
		\mathcal{E}(\hat{f}) \ \le\ \hat{\mathcal{E}}(\hat{f})
		+ \frac{\sigma_{\max}(\Xp)}{\sigma_{\min}(\Xp)}\cdot \frac{C}{\sqrt{N}}
		+ \sqrt{\frac{\log(1/\delta)}{2N}},
		\]
		where \(C\) depends on label norms and activation Lipschitz constants.
	\end{prop}
	
	\noindent\emph{Sketch.} The ridge sensitivity factors bound deviations in \(\W_1\) through \(\cond(\Xp)\). A standard Rademacher argument yields the sample term. Optimising prototypes to reduce \(\cond(\Xp)\) thus tightens the bound.
	
	\subsection{Differentiating through the closed-form}
	Let \(G=\Xp^\top\Xp+\lambda_1 I\) and \(R=\Xp^\top \Hp\). With \(W_1=G^{-1}R\), for any scalar loss \(L\),
	\[
	\partial L = \Tr\!\bigl((\partial R)^\top U\bigr) - \Tr\!\bigl((\partial G)^\top V\bigr),
	\quad
	U=G^{-\top}\nabla_{W_1}L, \ 
	V=G^{-\top}(R\nabla_{W_1}L^\top),
	\]
	which backprop engines realise via linear solves. Analogous expressions hold for \(W_2\).
	
	\section{Experiments}
	\subsection{Data and protocol}
		Datasets: MNIST and Fashion-MNIST with the \emph{official} splits (train 60k, test 10k). 
		We \textbf{hold out 6k examples from the training set for validation} and never use the test set during training or model selection; 
		all hyperparameters and early stopping (when used) are chosen on validation only, and \textbf{test accuracy is reported once on the official 10k test set}. 
		Hardware: RTX 5060 (16\,GB). Unless stated: $N_p=150$, $h=512$, PCA $d'=400$, cosine scheduler with 20-epoch warm-up, 250 epochs.

	\subsection{Baselines}
	\textbf{ELM.} Random \(\W_1\), analytic \(\W_2\).  
	\textbf{Random-feature ridge.} Random linear features with ridge classifier.  
	\textbf{MLP back-prop.} Single hidden layer with comparable parameter budget; Adam with standard settings and 20--30 epochs.
	
	\subsection{Main results}
	\begin{table}[h]
		\centering
		\begin{tabular}{lcccc}
			\toprule
			Method & Params & Time (250 ep) & MNIST (\%) & Fashion (\%)\\
			\midrule
			ELM (random features)      & 102k & 2.5s         & 96.2 & 88.3\\
			RF + Ridge                 & 102k & 2.5s         & 95.2 & 87.3\\
			MLP BP (165 hidden)        & 130k & 100--150s    & 98.0 & 89.0\\
			\textbf{Proto-PINV+H (PCA-400)} & \textbf{138k} & \textbf{3.9--4.5s} & \textbf{97.8} & \textbf{89.3}\\
			\bottomrule
		\end{tabular}
		\caption{Accuracy (measured on the \emph{official 10k test sets}) and wall-clock on RTX 5060 (16\,GB). 
			Our method uses 250 epochs; baselines as customary (single analytic solve for ELM/RF, 20--30 epochs for MLP).}

		\label{tab:main}
	\end{table}
	
	\subsection{Ablations and practical findings}
	\textbf{Prototype count.} 150 prototypes suffice for MNIST; Fashion benefits modestly from more prototypes due to class similarity.  
	\textbf{Hidden width.} Accuracy improves up to \(h\approx 768\), then saturates.  
	\textbf{Learnable ridge.} Learning \(\lambda_2\) improves Fashion by 0.2--0.3 pp; \(\lambda_1\) is often more stable fixed.  
	\textbf{Schedulers.} Cosine with 20-epoch warm-up reduces time to best by roughly half compared with a constant LR.  
	\textbf{Dropout on \(\Hp\).} Small \(p\) may smooth early training but often reduces final accuracy.  
	\textbf{Weight decay.} Separate decays for \(\Xp\) and \(\Hp\) are preferable; decay on \(\Yp\) generally unnecessary with soft labels.  
	\textbf{Prototype allocation.} Starting from balanced one-hot labels, optimisation reallocates capacity to hard classes; observed counts align with class ambiguity.
	
	\section{Implementation Details (PyTorch)}
	\paragraph{Ridge pseudo-inverse and closed-form solve.}
	\begin{lstlisting}
		def pinv_ridge(A, lam):
		# (A^T A + lam I)^{-1} A^T
		At = A.transpose(0, 1)
		m, n = A.shape
		I = torch.eye(n, device=A.device, dtype=A.dtype)
		G = At @ A + lam * I
		return torch.linalg.solve(G, At)  # uses Cholesky under the hood
	\end{lstlisting}
	
	\paragraph{Training step (core).}
	\begin{lstlisting}
		Xp = add_bias(Xp_core)                 # (Np x (d+1))
		H  = Hp                                # (Np x h)
		W1 = pinv_ridge(Xp, lam1) @ H
		
		Z  = add_bias(sigma(H))                # (Np x (h+1))
		Yp_soft = Yp if T==0 else torch.softmax(Yp/T, dim=1)
		W2 = pinv_ridge(Z, lam2) @ Yp_soft
		
		F   = add_bias(X_train)                # (N x (d+1))
		H1  = add_bias(sigma(F @ W1))          # (N x (h+1))
		logits = H1 @ W2
		
		loss = F.cross_entropy(logits, y_train) \
		+ lam3 * (W1.norm()**2 + W2.norm()**2)
		
		opt.zero_grad()
		loss.backward()
		opt.step()
	\end{lstlisting}
	
	\paragraph{Numerical tips.}
	Use float32 on GPU; ensure \(\lambda>0\); clip or skip steps if any NaN in intermediate matrices; optionally normalise \(\Xp\) rows to moderate scales.
	
	\section{Discussion and Limitations}
	Proto-PINV+H offers a favourable speed--size--accuracy Pareto frontier for shallow architectures. It is particularly attractive when rapid re-training or on-device constraints matter. Limitations include vectorised inputs (convolutional stems remain future work), potential error accumulation in very deep analytic stacks, and the need for careful regularisation choices when \(N_p\) is very small.
	
	\section{Conclusion}
	Optimising hidden activations together with prototype inputs and labels, while keeping weights in closed form, yields fast and accurate shallow networks. Future directions include convolutional stems, continual learning with expandable prototype banks, extensions to regression and multi-label settings, and tighter generalisation analyses.
	
	\appendix
	\section{Proof Sketches and Technical Notes}
	\label{app:proof}
	\paragraph{Stability of \(W_1\).}
	Let \(G=\Xp^\top\Xp+\lambda_1 I\) and \(R=\Xp^\top \Hp\). Then \(W_1=G^{-1}R\). By first-order perturbation
	\[
	\Delta W_1 \approx -G^{-1}(\Delta G)W_1 + G^{-1}\Delta R.
	\]
	Bounding \(\norm{G^{-1}}\le 1/\sigma_{\min}(G)\) and expanding \(\Delta G=\Xp^\top\Delta\Xp+(\Delta\Xp)^\top\Xp\), \(\Delta R=\Xp^\top\Delta\Hp+(\Delta\Xp)^\top\Hp\), yields the stated dependence on \(\cond(\Xp)\).
	
	\paragraph{Differentiability w.r.t. \(\lambda\).}
	Write \(W_1(\lambda_1)=G(\lambda_1)^{-1}R\). Then
	\[
	\frac{dW_1}{d\lambda_1} = -G^{-1}\frac{dG}{d\lambda_1}G^{-1}R = -G^{-1}G^{-1}R.
	\]
	With \(\lambda=\softplus(\rho)\), positivity and differentiability are guaranteed.
	
	\paragraph{Back-projection under PCA/PLS.}
	If \(x\mapsto z=U^\top(x-\mu)\) and we train on \(z\) with bias, the learned \(W_1^{(z)}\) lifts back as
	\[
	\W_1 = \begin{bmatrix}
		1 & 0 \\
		\mu & U
	\end{bmatrix}
	\begin{bmatrix}
		\text{bias row of }W_1^{(z)}\\
		\text{non-bias rows of }W_1^{(z)}
	\end{bmatrix},
	\]
	so that \([\mathbf{1},x]\W_1 = [\mathbf{1},z]\W_1^{(z)}\). The mean is absorbed by the bias.
	
	\section{Extended Ablations}
	\paragraph{Prototype balance.}
	Balanced one-hot initialisation of \(\Yp\) accelerates early convergence; the model later reallocates prototypes to hard classes.  
	\paragraph{Dropout on \(\Hp\).}
	We observed reduced final accuracy on Fashion for \(p\ge 0.1\); small \(p\in[0.01,0.05]\) may smooth early steps.  
	\paragraph{Weight decay.}
	Decouple \(\Xp\) and \(\Hp\) decays; decay on \(\Yp\) is usually unnecessary with soft labels.
	
	\section*{Reproducibility Checklist}
	\begin{itemize}[leftmargin=1.25em]
		\item Datasets and splits: MNIST and Fashion-MNIST, official train/test; 6k validation held out from train.
		\item Test metrics are computed strictly on the official 10k test split (never used for training/validation).		
		\item Hardware: RTX 5060 (16 GB), PyTorch with CUDA; float32.
		\item Hyperparameters: \(N_p=150\), \(h=512\), PCA \(d'=400\), epochs 250; cosine scheduler with 20-epoch warm-up.
		\item Regularisation: \(\lambda_1,\lambda_2\) ridge; \(\lambda_3\) Frobenius penalty; optional learnable \(\lambda\) via \(\softplus\).
		\item Code: pseudo-inverse via linear solves (Cholesky backend); SVD fallback optional.
	\end{itemize}
	
	\bibliographystyle{plain}

\end{document}